\documentclass{article}

\usepackage{arxiv}

\usepackage[utf8]{inputenc} 
\usepackage[T1]{fontenc}    
\usepackage{hyperref}       
\usepackage{url}            
\usepackage{booktabs}       
\usepackage{amsfonts}       
\usepackage{nicefrac}       
\usepackage{microtype}      
\usepackage{lipsum}		
\usepackage{graphicx}
\usepackage{natbib}
\usepackage{doi}
\usepackage{subfig}
\usepackage{algorithm}
\usepackage{algpseudocode}

\title{Public Parking Spot Detection And Geo-localization Using Transfer Learning}

\author{ \href{https://orcid.org/0000-0002-9191-0565}{\includegraphics[scale=0.06]{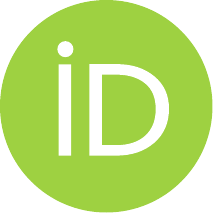}\hspace{1mm}Moseli Mots'oehli} \\
	Department of Information and Computer Science\\
	University of Hawai'i At Manoa\\
	Honolulu, HI 96822 \\
	\texttt{moselim@hawaii.edu} \\
	\And
	\href{https://orcid.org/0000-0001-5626-6172}{\includegraphics[scale=0.06]{orcid.pdf}\hspace{1mm}Yao Chao Yang} \\
	Department of Actuarial Science\\
	University Of Pretoria\\
	Pretoria, 0002 \\
	\texttt{u12082602@tuks.co.za} \\
}

\renewcommand{\shorttitle}{Parking Spot Detection Using ML}

\hypersetup{
pdftitle={A template for the arxiv style},
pdfsubject={q-bio.NC, q-bio.QM},
pdfauthor={Moseli Mots'oehli, Yao Chao Yang},
pdfkeywords={Parking Detection, Geolocalization, Transfer Learning, Haversine Distance},
}

\graphicspath{ {./images/} }

\begin{document}
\maketitle

\begin{abstract}
In cities around the world, locating public parking lots with vacant parking spots is a major problem, costing commuters time and adding to traffic congestion. This work illustrates how a dataset of geo-tagged images from a mobile phone camera, can be used in navigating to the most convenient public parking lot in Johannesburg with an available parking space, detected by a neural network powered-public camera. The images are used to fine-tune a Detectron2 model pre-trained on the ImageNet dataset to demonstrate detection and segmentation of vacant parking spots. We use the parking lot's corresponding longitude and latitude coordinates to recommend the most convenient parking lot to the driver based on their Haversine distance to a lot, and the number of open parking spots. The VGG Image Annotation (VIA) tool was used to annotate images with polygon outlines of the four different types of objects of interest: cars, open parking spots, people, and car number plates. We detect and segment number plates to ensure they can be occluded in production for car registration anonymity, and driver privacy. Intersection over union cover scores of 89\% and 82\%  on cars and parking spaces, respectively, were achieved on the test set. This work has the potential to help reduce the amount of time commuters spend searching for free public parking, hence easing traffic congestion in and around shopping complexes.

\end{abstract}

\section{Introduction}\label{sec:introduction}
With the world trend of the population moving from rural to urban cities, the densities are increasing rapidly in large cities like Johannesburg, South Africa. The result of this urbanization is an increase in vehicles in the city, which leads to an increase in traffic congestion, air pollution, vehicle theft, and the risk of road accidents. These factors impact the economic activities in the city as well as the social and environmental aspects \cite{Hadi:VisionBasedPL19}. Another major problem in cities around the world is locating vacant parking spaces. This can cost commuters time and fuel, add to traffic congestion, accidents and aggressive driving \cite{Rosamaria:ParkSmart13,Ruben:OnStreetRandomundersampling21,Naveen:OpenAirParking22,Yan:RoadRage20}. The public parking spaces available on the street and at the shopping centers cannot be reserved in advance, and with limited space available, vehicles searching and stalling for the next available space are the cause of traffic. Furthermore, the vehicle emission level during the wait and circulation increases air pollution and is an environmental concern at a global level  \cite{Rosamaria:ParkSmart13,Ruben:OnStreetRandomundersampling21,Naveen:OpenAirParking22}. The continual search for available space is a frustrating task for drivers and the situation can be exacerbated during crowded peak-time hours and for new drivers, who may not be familiar with the surrounding area.

Most free public parking lots also lack the security features to deter theft or alert the driver when their car vacates a parking space they know it to be at, leading commuters to opt for paid, more secure and yet congested parking lots, leaving free access lots under-utilized. Many real-time parking management systems have been proposed as solutions to these issues with varying technological and social components. We hypothesize the ideal solution would optimize the limited lot space available \cite{Ruben:OnStreetRandomundersampling21} and fuel saving for drivers. These system are designed to identify a location with a free parking space, count the vacant spaces, track and report the changes to the driver in real-time with minimal latency. With the advancement of technology, streets and building are becoming more and more connected to the internet through the use of sensors and cameras. Some solutions in this domain utilize sensors, mounted cameras or drone-based camera systems, and machine learning methods, however, most of these do not address the issue of finding the best parking lot for a driver based on their current location and availability of parking.

In this paper, we illustrate how: by leveraging deep learning and the power of transfer learning, a small dataset of public parking lot images captured using mobile phone cameras in Johannesburg, South African, together with the parking lot's longitude and latitude coordinates information can be used to guide drivers to the most convenient parking lot based on their current location. This work has the potential to inspire both further research, and commercial applications of intelligent parking solutions to enhance and advance the parking lot management systems in South Africa. In the next section, we explore and critique existing literature in this domain.

\section{Related Work}\label{sec:Related Work}
The two popular input devices for capturing parking information are cameras and floor sensors. Ultrasonic Sensor based parking systems are normally installed in the middle or in front of each parking spot, and a network of them work to feed information to a central server. While technological progress has allowed for the deployment of cheaper and better sensors as explained in \cite{Jeffrey:WirelessParking14,Oludolapo:ScalableSmart21S}, multiple sensor installations remain significantly complex to install and costlier than current basic cameras \cite{Akkaynak:commercialCams2014}. Sensor technology for parking lot systems does not scale as well as cameras since each sensor typically only covers a few parking spots at a time. Camera-based solutions on the other hand have shown great promise due to advances in convolutional neural networks (CNNs) \cite{Dhuri:RealTimeVGG21,Hadi:VisionBasedPL19}, graphical processing units (GPUs), camera lenses, and the decreasing cost of such technology \cite{Naveen:OpenAirParking22,Hilal:IntelligentPark14,ShengFuu:VisionBased06,Imen:VPLD14}.

CNNs are commonly used for feature extraction in data containing spatial correlations \cite{Simonyan:VDCNN15,Alex:DeepCNN12}. Open-source image datasets such as ImageNet, CelebA, and MS-COCO sparked the explosive growth and success of CNNs in vision tasks \cite{Jia:ImageNet09,Lin:MSCOCO14,Liu:celebA14}. These large collections of data have allowed for building of very deep and complex neural networks with little risk of over-fitting \cite{Alex:DeepCNN12}. Authors of \cite{Al-Kharusi:IntelligentPark14} analysed an aerial view of a car park and developed a space detection model that determines whether a parking space is vacant or filled. In \cite{ShengFuu:VisionBased06} a  four camera parking lot system is developed by setting up each of the cameras around a building, combining RGB images and shadow detection for local robustness and better performance. In \cite{Hadi:VisionBasedPL19},the PKLot dataset \cite{Almeida:PKLot2015} samples under different weather conditions using a chromatic gradient analysis. They incorporated this weather is is analytically examined, and results used to compensate for weather condition changes in different parking spaces.

Authors of \cite{Ruben:OnStreetRandomundersampling21} examined the future parking occupancy problem under a traditional regression technique and a classification technique. By using random under-sampling, they overcome issues related to class imbalance, and produce results indicating classification outperforms the regression technique in future parking prediction. While we also classify different objects in an image, our work goes on to segment out the exact locations of objects such as people and car plates. In \cite{Naveen:OpenAirParking22}, the author applies a neural network to develop a real-time open-air intelligent parking system. While they train on all 24-hour light conditions and achieve great results, the dataset is limited to a single parking lot hence not allowing for adaptation in different landscape settings such as places with more shade during the day or lots with many trees. In \cite{Rosamaria:ParkSmart13}, an architectural framework for software, hardware integration, and operation of intelligent parking assistant systems is developed, and tested under simulation data similar to our approach this work. The results indicate that the intelligent parking assistant outperforms the conventional parking system.

In the South African context, the author of \cite{Nyambal:AutoParkCNN17} develops a CNN based system as we do, but with training data limited to the parking lots within WITS university main campus. In their approach, the problem is set up as a classification task, where each parking space is labelled empty or vacant. In real-time, this system would be incapable of tracking vehicle trajectories towards an open parking space since the neural network looks nowhere but between white parallel parking lines. In all works detailed so far, camera angle for images were ideal, the shot is taken from either directly in front or behind the cars, at a height that allows full visibility of all cars and parking spaces. As such, these methods do not address the use of single camera per lot, placed on buildings, at an angle to the cars, and parking spaces such that large portions of certain key objects are occluded by vehicles in neighbouring spots. We feel this is the more realistic case if one were to deploy such system at scale using existing surveillance cameras in shopping complexes. CNN's are also used in \cite{Xu:ParkingFisheye20}, and in particular: YOLOV3, an architecture much similar to our implementation than any of the previously stated work. The authors of \cite{Xu:ParkingFisheye20} opt for a car mounted 4 way camera for data collection as they drive around parking lots.

Most work in this domain has been on a single parking lot, and not a network of parking lots. A multi-lot system would enable parking spot recommendations for drivers based on availability of open spaces, and proximity to the parking destination. The task of recommending an ideal location based on distance has been tackled in other domains where proximity is vital, such as nearest emergency center recommendation systems \cite{Basyir:NearestEmergency17,Rezania:TempShelter14,Berker:AreaFPGA21,Hagar:ShortestPath16}. The dominant distance measure used in these works is the Haversine formula. While it has been shown to produce errors of up to 0.5\% of the distance being measured in Kilometers, it has been deemed a reliable measure of distance between two points on earth.

\section{Data Annotation}\label{sec:Data}
To be able to first detect objects of interest, segment, and classify each of them as either a car, number plate, parking spot or person, we hand label, and annotate a small but representative sample of phone camera images from the overall dataset. These hand labelled images are used for fine-tuning parameters of a pre-trained Detectron model as will be explained in the next section. We use the open sourced VGG Annotator to input polygon coordinates of the different objects of interest, and their classes to form the fine-tuning dataset. In doing this, it was necessary to use images that range from a perfectly visible parking spot, a partially occluded parking spot, and bad camera angles showing only a fraction of the parking space. We do the same for car, and number plate visibility. This is done to include as many edge cases as possible that could be encountered in the test environment. We put a strong emphasis on number plate detection to be able to crop out or blind all detected number plates in the visualization, and deployment of this work. Below are examples of annotated images showing the polylines around the different objects of interest.

\begin{figure}%
    \centering
    \subfloat[\centering Annotation with multiple open spots]{{\includegraphics[width=6cm]{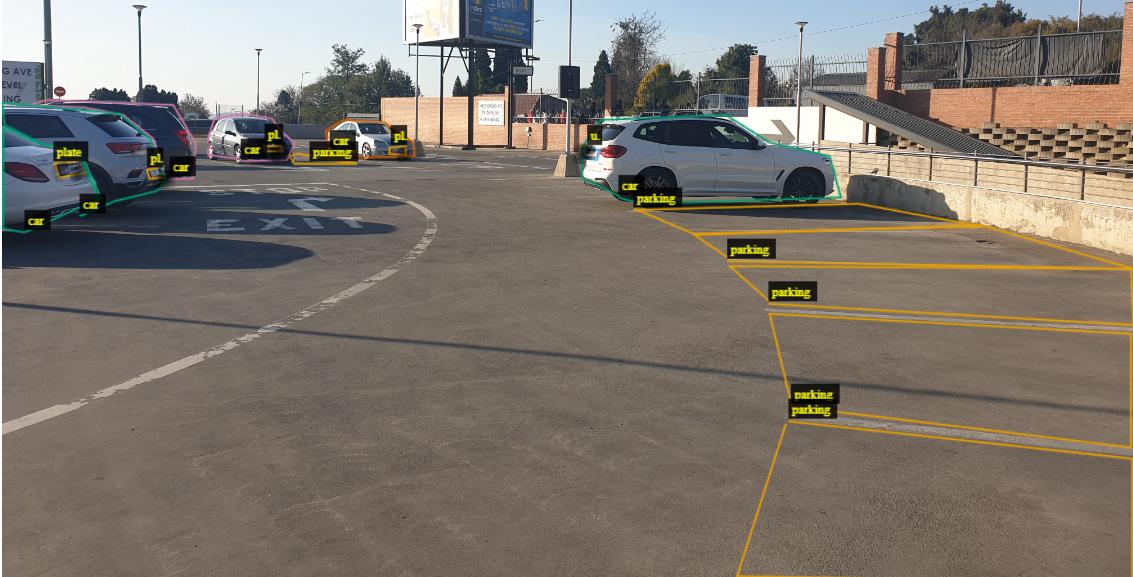} }}%
    \subfloat[\centering Annotation with roadside parking open spots]{{\includegraphics[width=6cm]{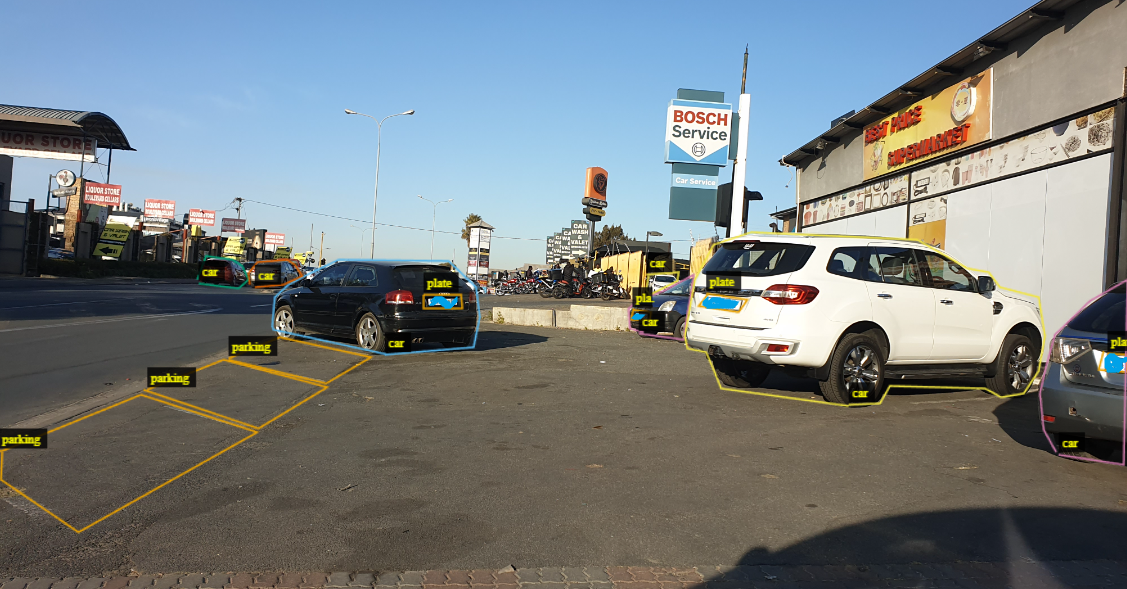} }}%
    \qquad
    \subfloat[\centering Annotation in dimmer evening light]{{\includegraphics[width=6cm]{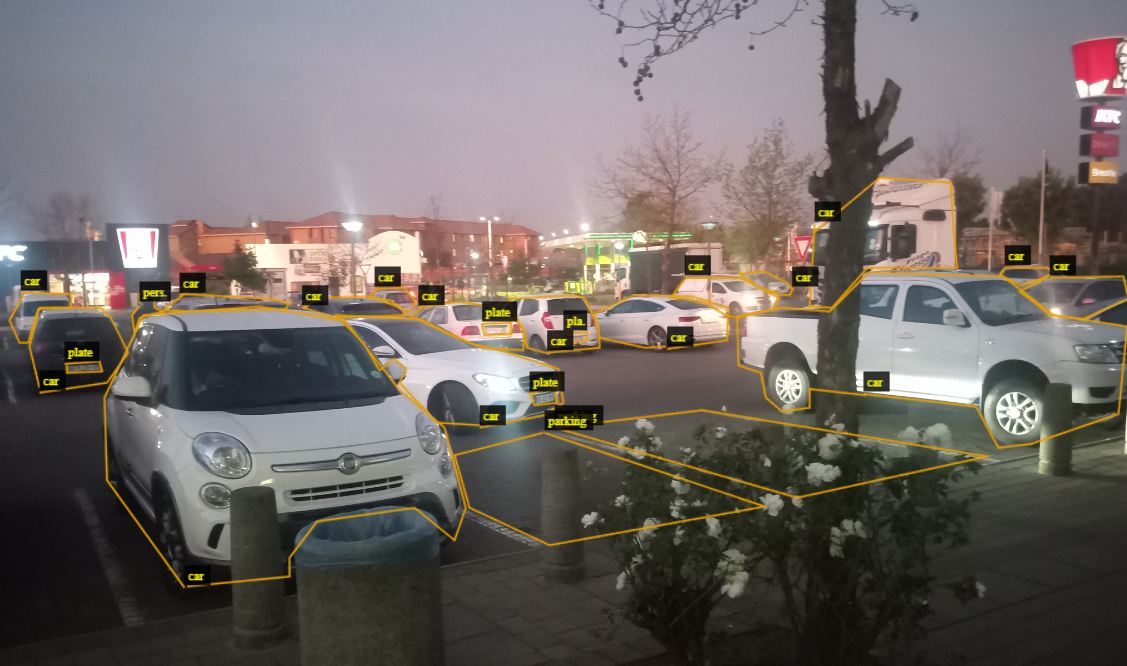} }}%
    \subfloat[\centering Annotation showing a person does not take up a parking space ]{{\includegraphics[width=6cm]{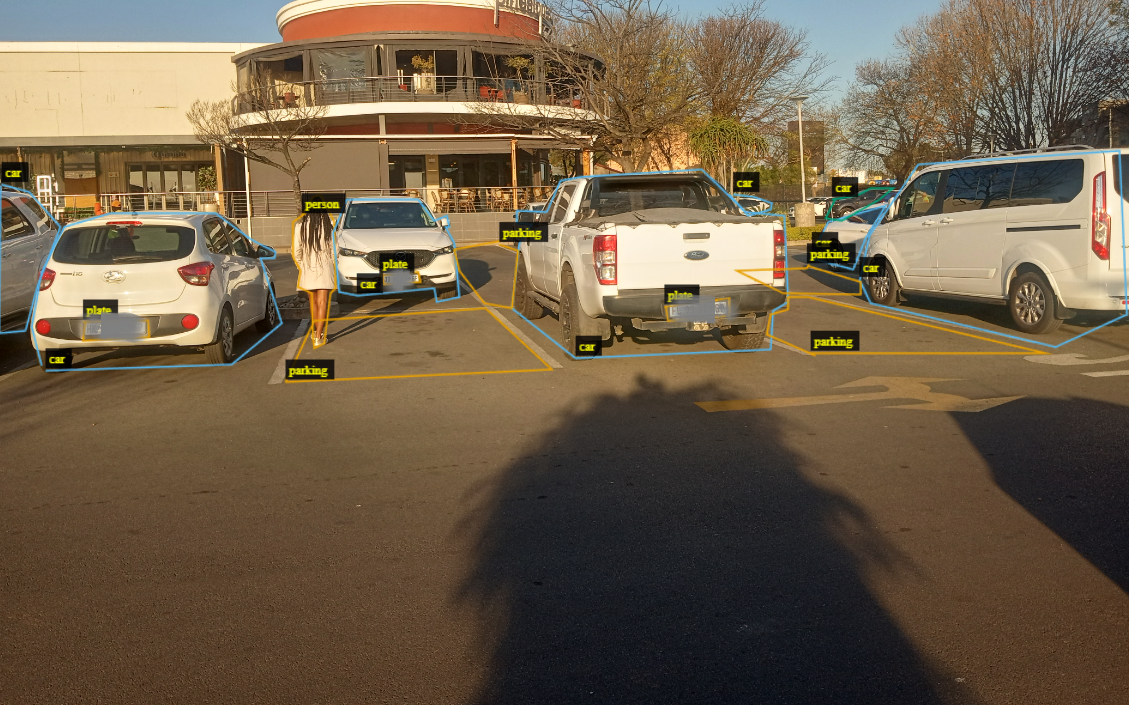} }}%
    \caption{The sample images above show the different scenarios that may be encountered in parking lots, and how the the annotations will look like to the Detectron2 model.}%
    \label{fig:example annotated images}%
\end{figure}

The dataset is compiled so that every batch of images from a specific public shopping centre parking lot is tagged with the longitude and latitude coordinates, which are potentially useful when deploying a mobile application powered by this parking spot detector. The coordinates point to each parking lot, and not necessarily the exact location of a rectangular parking space. Table \ref{Fine-tuning dataset} shows the number of images as well as the geo-location for the different free parking lots contained in the overall dataset.

\begin{table}
\centering
\caption{Number of images per parking lot as well as geo-Location (accurate to 4 decimal places).}\label{Fine-tuning dataset}
\begin{tabular}{|l|l|l|}
\hline
{\bfseries Parking Lot, and \#} & {\bfseries Geo-location} & {\bfseries Number of Images}\\
\hline
\hline
Brentwood Mall(1) &  $(-26.1189, 28.2804)$ & 3\\
Engen Morningside service(2) &  $(-26.0709, 28.0644)$ & 3\\
Intercare fourways(3) & $(-26.0158, 28.0064)$ & 25\\
Morning Glen Mall(4) & $(-26.0659, 28.0736)$ & 9\\
Pineslope(5) & $(-26.0209, 28.0139)$ & 17\\
Rivonia Junction Centre(6) & $(-26.0597, 28.0600)$ & 12\\
Best price supermarket Edenvale(7) & $(-26.0540, 28.0552)$ & 7\\
\hline
\textbf{Total} & - & \textbf{76}\\
\hline
\end{tabular}
\end{table}

\section{Methods}\label{sec:Methods}
\begin{figure}
\centering
\includegraphics[scale=0.65]{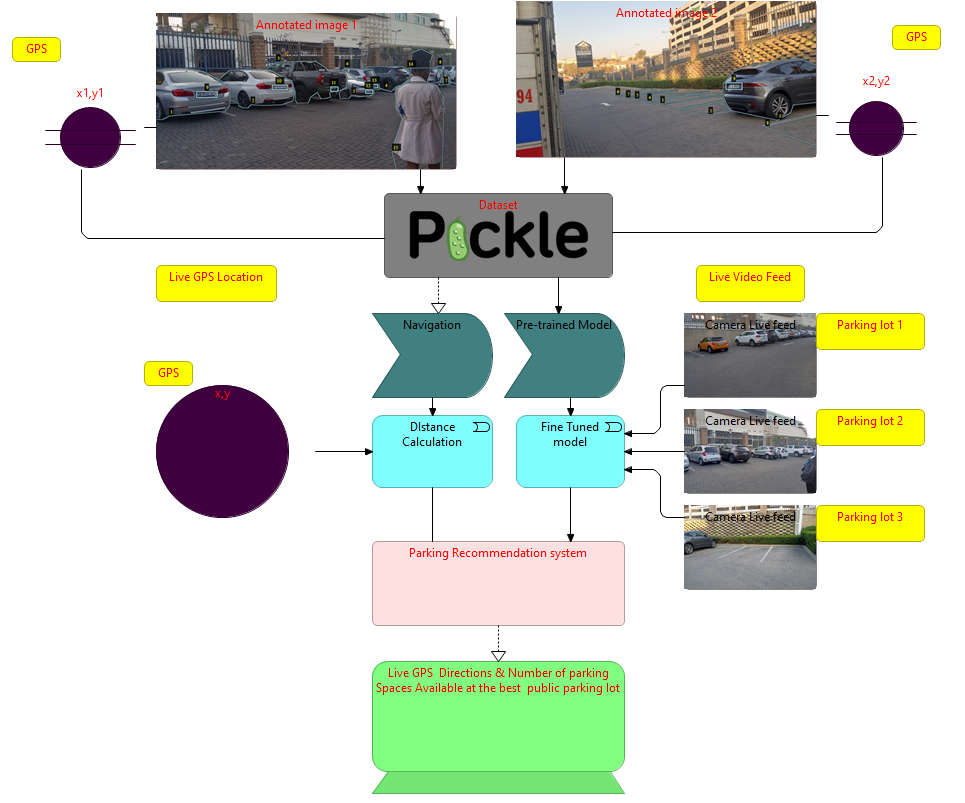}
\caption{The proposed system for public parking space detection and navigation recommendation. We use the longitude(x) and latitude(y) coordinates of a driver to suggest the best parking lot in terms of distance and number of available parking spaces.} \label{fig:linear_system}
\end{figure}

The overall system is iterative in nature:  first, we use a pre-trained CNN-based architecture to detect, and segment objects of interest. We then use these on a live video feed to suggest parking lots with the best location, and parking space count combination. In production, the user is asked through a mobile application whether the suggested parking lot, and space were both available and most convenient. We use this feedback and video frame prediction that led to the suggestion as additional training data for our detection and segmentation model. The response will also be used in the future to train an ensemble model with the user's GPS location and all parking lots as inputs, and the closest lot ID as output. This could potentially replace or supplement the distance calculation we perform as explained in \ref{subsec: Distance estimation}. Figure \ref{fig:linear_system} depicts The proposed system. Below we detail how the Detectron2 model is used for parking spot detection, and how the distance to a parking lot is calculated.

\subsection{Detectron2}\label{subsec:Detectron2}
\begin{figure}
\centering
\includegraphics[scale=0.85]{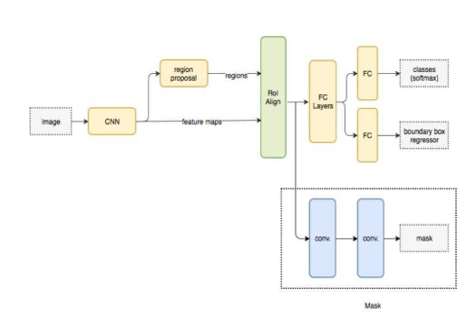}
\caption{The backbone of Detectron2 is an R-CNN model that uses learned image features to perform three tasks, object classification, bounding box regression and segmentation mask prediction.} \label{fig:detectron2}
\end{figure}

Detectron2 written in Pytorch, is a collection of re-implementations of state-of-the-art object-detection algorithms including Mask R-CNN~\cite{He:maskrcnn17} for detection and segmentation. Mask R-CNN trains a single CNN for pixel-wise segmentation, classification, and bounding box regression. As depicted in Figure ~\ref{fig:detectron2}, a pre-defined number of regions of interest (ROIs) are proposed, of adjacent pixels similar in color, texture, or intensity from the features learned by a stack of convolutional layers. The network is then trained to minimize the classification and segmentation losses of the best ROIs. Once ROIs with very high probabilities of containing objects are found, a fully connected layer is added to predict the x, and y coordinates of a rectangular region that most tightly encloses the objects with high confidence. Given a new image, the trained model produces a list of detected objects, each with the following information: (1) the predicted class (car, parking, person, or plate); (2) a bounding box that represents the smallest rectangular region that completely contains the detected object; (3) a pixel-wise segmentation mask outlining the object; (4) an object-level predicted class confidence score.

\subsection{Unique Parking Spot Identification}\label{subsec: Unique parkking}
Since cameras send a continuous stream of video to the server for inference, counting the number of available spots at any one time in a parking lot depends on the number of frames in between successive inferences. Running inference on each frame of the video also affects the speed of the overall system, and so some engineering is required to ensure low latency while maintaining high detection and segmentation accuracy. To be able to tell that two detected parking spaces between successive frames are exactly the same parking spot, and so should not be double counted, we use the pixel locations of the bounding box around the detected parking space, and calculate the Intersection over union(IoU) of the two rectangular areas:

\begin{equation}
IoU = \frac{Area overlap}{Area union} 
\end{equation}\label{eq:IoU}

 The camera angle and distance to the objects, in the real-life setting will be fixed, and so a high IoU indicates the two detected parking spots within a given time interval or number of frames are the same parking spot. This method helps avoid the need to assign unique IDs to each parking space during data collection and model training.

\subsection{Distance Estimation and Navigation}\label{subsec: Distance estimation}
Once a vehicle is in search for parking through the application with geo-location permission allowed, we calculate the distances of all parking lots with a detected available parking spot. The ideal recommendation would consist of the closest parking lot, with the most parking spots available to choose from, and so the optimization problem can be stated as follows:

Given we have functional cameras at $N$ public parking lots, each with geo-location $(x_{i},y_{i})$ and $m_{i}$ detected available parking spots, where $m_{i} \ge 1$, a user/driver at location $(x^*,y^*)$ wants a combination of the smallest distance and largest number of parking spots, 

\begin{equation}
\min_{i} \quad ||(x^*,y^*) -(x_{i},y_{i})||, \quad for \quad i=1,2,3...,N
\end{equation}\label{eq:min_distance}

and 

\begin{equation}
\max_{i} \quad m_{i}, \quad for \quad i=1,2,3...,N
\end{equation}\label{eq:max_m}

The distance measure $||\cdot||$ in the case of longitude and latitude points on a sphere is the haversine distance, that has been shown a very high level of accuracy in past applications \cite{Rezania:TempShelter14,Basyir:NearestEmergency17,Berker:AreaFPGA21}. The haversine distance $d_{i}$ of each parking lot $i$ from the driver's current location $*$ is given by:

\begin{equation}
d_{i} = 2R\arcsin{\sqrt{\sin^2{\frac{\Delta_{x}}{2}} + \cos{x^*}\cos{x_{i}}\sin^2{\frac{\Delta_{y}}{2}}}},
\end{equation}\label{eq:haversine}

where $\Delta_{x} = x_{i}-x^*$ and $\Delta_{y} = y_{i}-y^*$  for $i = 1,2,3,...,N$

Using the haversine  distance, and inverting the number of spots available, equations \ref{eq:haversine} and  \ref{eq:max_m} can be posed as a combined minimization problem with objective function:

\begin{equation}
\min_{i} \quad \alpha d_{i} + \frac{1-\alpha}{m_{i}}, \quad for \quad i=1,2,3...,N
\end{equation}\label{eq:objective function}

with $0 \le \alpha \le 1$ set to ensure distance and the number of parking spots are approximately equally important. The $\alpha$ parameter can also be learned and adjusted based on each drivers preference whenever a driver opts for a parking lot other than the recommended lot.This is left for future research. Both parts of the objective function can be computed in $\mathcal{O}(n)$ time, so we can write an algorithm solve the optimization problem in $\mathcal{O}(n)$ time and $\mathcal{O}(1)$ space as shown in algorithm \ref{alg:Distance_and_Spot}.

\begin{algorithm}
\caption{Best parking spot recommendation algorithm}\label{alg:Distance_and_Spot}
\begin{algorithmic}
\Require $N \ge 1$
\State $i \gets 1$
\State $Best_i \gets 1$
\State $Best \gets \infty$
\While{$i \le N$}
\If{$m_{i} == 0$}  \Comment{skip full parking lots}
\State $pass$
\Else
\State $d \gets \alpha \times ||(x^*,y^*)-(x_{i},y_{i})||$
\State $m \gets \frac{1-\alpha}{m_{i}}$
\If{$d+m \le Best$}
    \State $Best \gets d+m$
    \State $Best_i \gets i$
\Else
    \State $pass$
\EndIf
\EndIf
\EndWhile
\State $\textbf{Recommend parking lot} \quad Best_i$
\end{algorithmic}
\end{algorithm}

\section{Experimental Setup}\label{sec:Experiments}
As can be seen from the example annotations on section \ref{sec:Data}, hand labelling and segmenting cars, parking spaces and number plates requires exponentially more time than it takes to acquire images of parking lots. To make up for our small dataset of images, we make use of transfer learning and only learn high level features. The Detectron2 model we train is written in Pytorch, and is trained on a cluster node utilizing 32 Gigabytes of RAM and 2 NV-RTX2080Ti GPUs. The annotations are converted to COCO Instance Segmentation style, and are stored in a JSON file. We use a 70:30 data train, test split, a base learning rate of $0.0004$  with decay, and train for 3000 iteration. Random Flip, and shortest edge resizing  transformations are applied as images go through the generator.
For training, only the fully connected layer if trained and the convolutional layers are frozen. The original model was pre-trained on the ImageNet dataset with 10 classes, so we change the output layer to 4 classes corresponding to the 4 object types we are interested in. 

Since the training images were captured using standard phone cameras, and not actually mounted surveillance cameras at the parking lots, we use the test data images as well as the coordinates of each image's parking to simulate parking lot recommendations based on 5 randomly picked locations a driver could be at in Johannesburg. In this setting, each image in the test set belongs to exactly one parking lot, and when the first simulated driver location is presented, the system then returns a list of possible parking lots, ranked from best to worst based on the objective function in \ref{eq:objective function}. We do not demonstrate identification of unique parking spots between video frames since we only show results on captured images and not video feed from a mounted live camera. This is left for further research as it requires financial investment in buying camera hardware or permission to access surveillance feed from existing shopping complex cameras.

\begin{table}
\centering
\caption{Distribution of instances among all 4 categories.}\label{tab:Training_dataset_class_instances}
\begin{tabular}{|l|l|l|}
\hline
{\bfseries category} & {\bfseries \#instances} & $\%$ \\
\hline
\hline
car &  298 & 44\\
parking &  198 & 29\\
person & 32 & 5\\
Number plate & 153 & 22\\
\hline
\textbf{Total} & \textbf{681} & \textbf{100}\\
\hline
\end{tabular}
\end{table}

Table \ref{tab:Training_dataset_class_instances} shows the number of instances of each class in the training dataset.

\section{Results and Discussion}\label{sec:Results and Discussion}
In this section, Detectron2 test results based on detection, segmentation and classification are presented, followed by the simulated driver recommendations in 5 different locations looking for a convenient free parking spot. We will then show how varying $alpha$ affects the recommended parking lot from each starting point.

\begin{figure}%
    \centering
    \subfloat[\centering Example model prediction with 1 parking spot detected in a lineup of cars]{{\includegraphics[width=6cm]{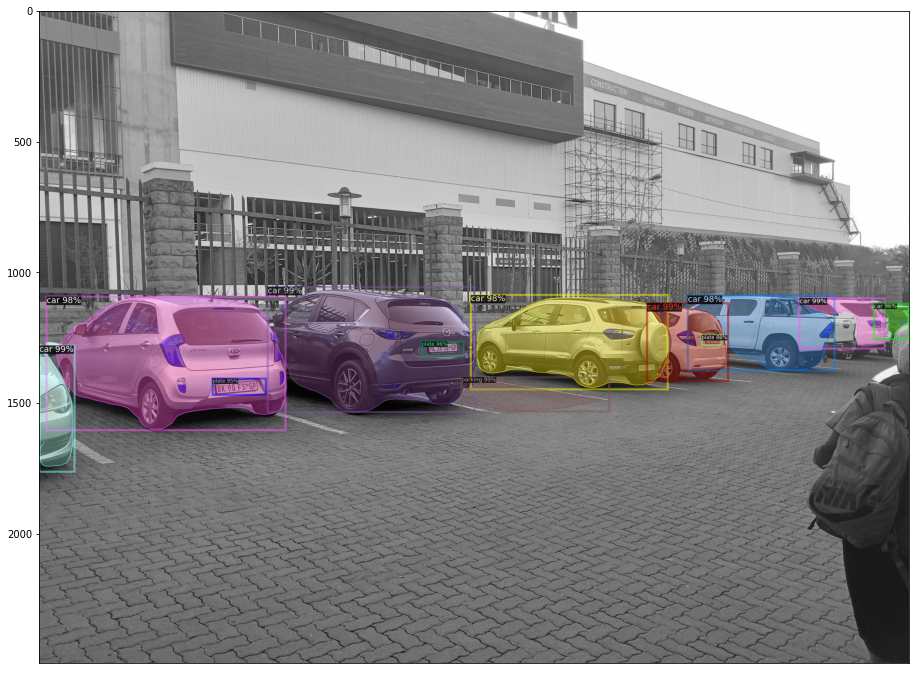} }}%
    \subfloat[\centering The model is able to detect cars and parking even in different parking orientations 2]{{\includegraphics[width=6cm]{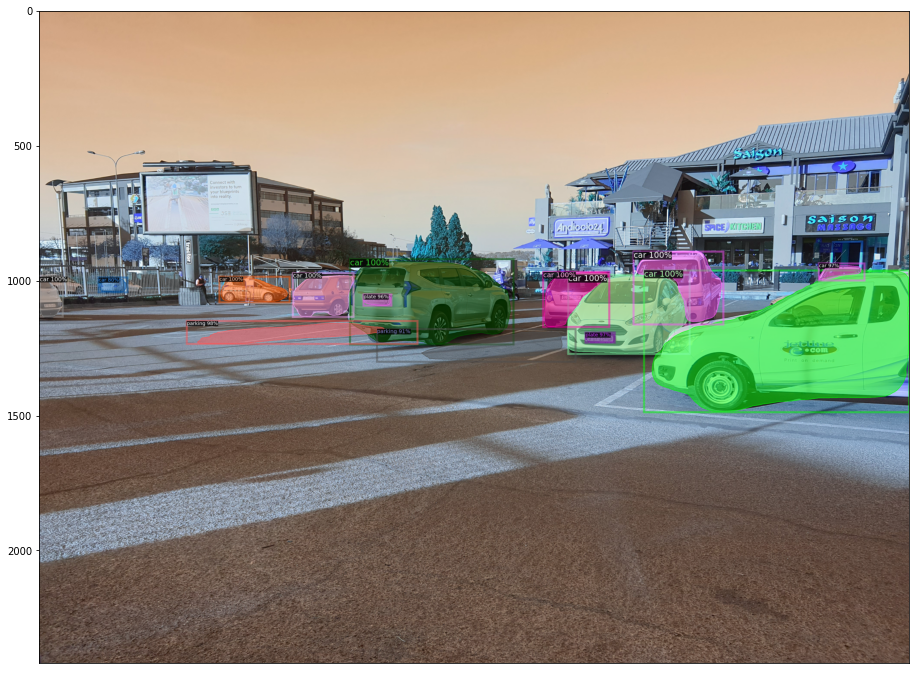} }}%
    \qquad
    \subfloat[\centering OOD prediction 1 showing more errors on detecting parking ]{{\includegraphics[width=6cm]{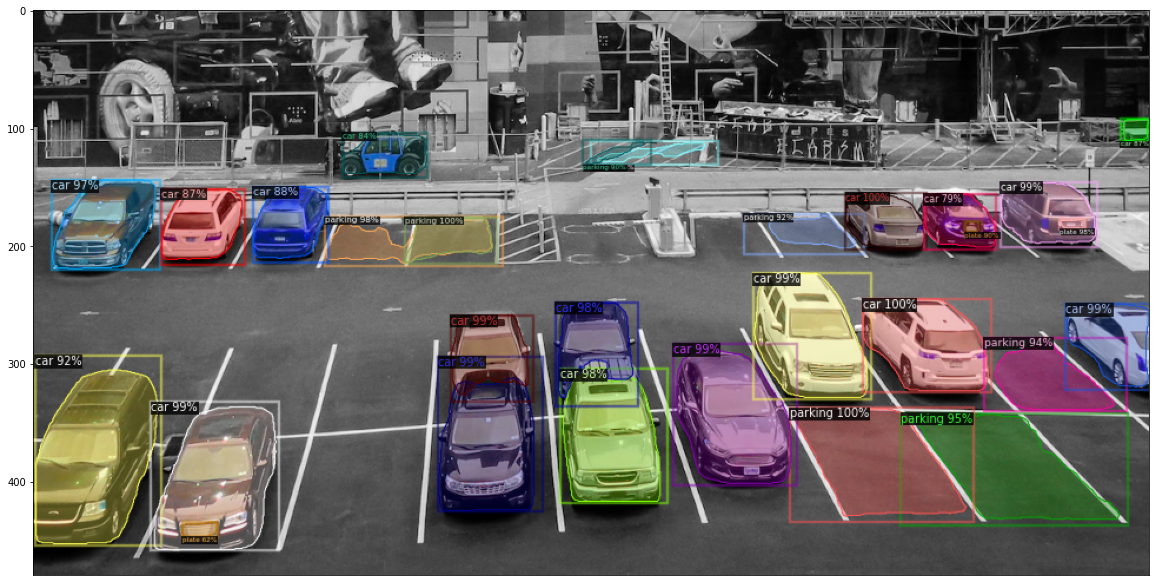} }}%
    \subfloat[\centering  OOD prediction 2 showing more errors on detecting parking]{{\includegraphics[width=6cm]{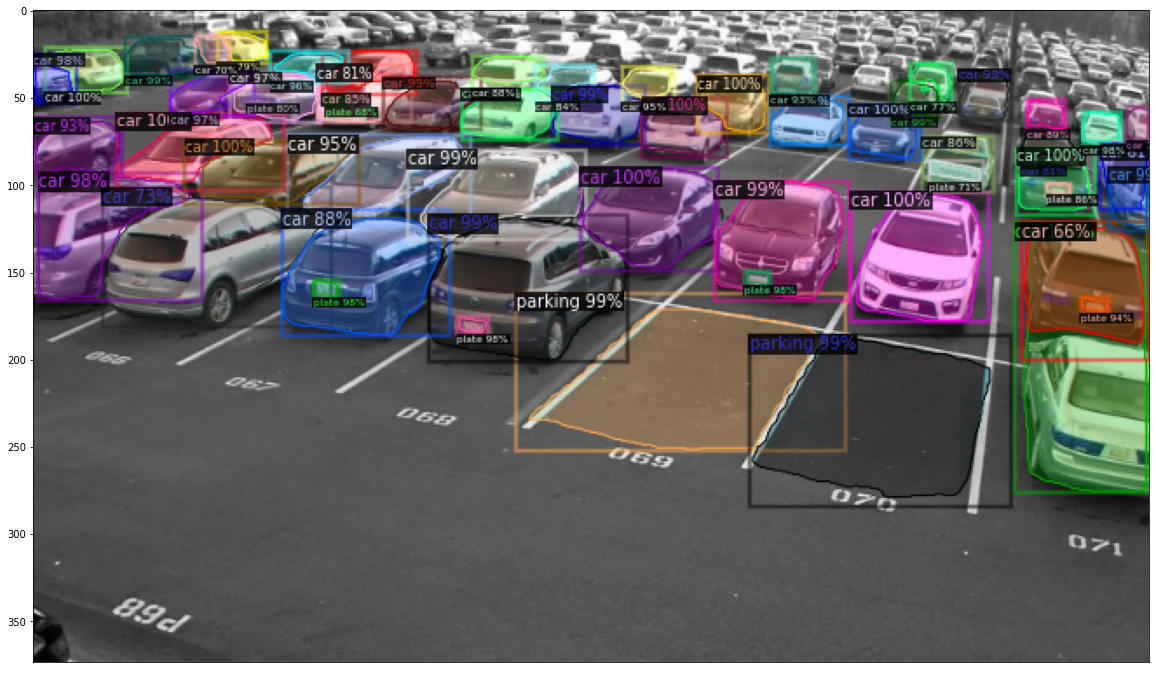} }}%
    \caption{Top right and Left: Sample predictions of the test data. Bottom right and Left: Sample predictions on Out Of Distribution (OOD) images taken in USA.}%
    \label{fig:example predictions images}%
\end{figure}

\subsection{Detection, Segmentation and Classification}\label{subsec: Detectron2_results}
\begin{figure}
\centering
\includegraphics[scale=0.5]{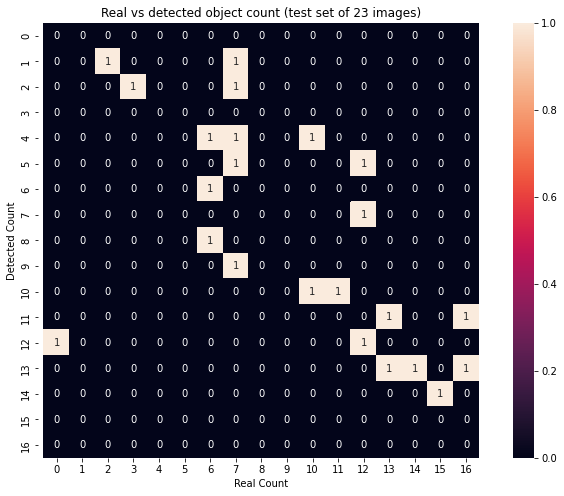}
\caption{Detected vs real object counts of categories: parking, car, person or number plate.} \label{fig:detection_heatmap}
\end{figure}

Evaluating detection and segmentation, We look at the number of detected objects verses the ground truth number of objects in each image. While this isn't a very good measure for how well a model can pin point exactly where objects of interest are, it is a good sanity check.  Figure \ref{fig:detection_heatmap} shows the detection count confusion matrix heat-map on the test set. In 17\% of the test images, the model is able to detect the correct number of objects of interest, and in 70\% of cases misses not more than 3 objects in an image. 

IoU scores were computed to evaluate how well the model is able to assign image pixels to an object of the correct class. A high IoU score (near 100) means the model is able to draw out the exact perimeter of a parking space or car. The table below contains IoU scores for the different classes on the test set. The high IoU scores for cars and parking lots can also be seen when predicted segmentation masks are overlayed on the input image as can be seen in figure \ref{fig:example predictions images}. In the same figure, (c) the model perfectly draws out the area of the parking lots it was able to detect even though this is an image not taken from Johannesburg, or similar angle as all the images in the training set. Performance was highest on cars, followed by number plates although they are much smaller than parking spaces. This could be due to the fact that number plate detection and segmentation is bounded within areas containing cars, but parking spaces can be anywhere. Both the training and test data contain relatively few examples of people, and in most cases the people are in the background, occluded by cars, or are so far it is hard even for a human observer to tell there is a person in the image. The results we obtain for cars, parking and plate number are satisfactory that a prototype mobile application would be functional with the currently trained model.

The model was used on the test set, together with GPS coordinates to demonstrate a simulated usage of the complete system below.

\begin{table}
\centering
\caption{IoU results on the test set split by all 4 categories of interest.}\label{tab:IoU_scores}
\begin{tabular}{|l|l|l|l|}
\hline
{\bfseries category} & {\bfseries \#instances} & {\bfseries Mean IoU} & {\bfseries Median IoU} \\
\hline
\hline
car &  119 & 89\% & 92\%\\
parking &  92 & 82\% & 83\%\\
person & 13 & 67\% & 69\%\\
Number plate & 66 & 85\% & 85\%\\
\hline
\textbf{Total} & \textbf{324} & \textbf{80.8\%} & \textbf{82.3\%
}\\
\hline
\end{tabular}
\end{table}

\begin{table}
\centering
\caption{Distance (in Km) and available parking spots. The closest parking lot from each starting point is in bold}\label{tab:distance_and_parking}
\begin{tabular}{|l|l|l|l|l|l|l|}
\hline
{\bfseries Lot} & {\bfseries Bushhill} & {\bfseries Waterval Ct} & {\bfseries Dobsonville} & {\bfseries Germiston \textbf{S}} & {\bfseries Eldoraigne} & {\bfseries Open spots}\\
\hline
\hline
1 & 35.1280	& 21.3675 &	42.2779	& 15.8420 & 34.2195	& 3\\
2 & 13.3258 & 7.7578 & \textbf{25.1326} & \textbf{19.8164} & 28.1739 & 5\\
3 & \textbf{10.3292} & 9.7322 & 25.9869 & 28.1369 & \textbf{25.4353} & 8\\
4 &	14.2915 & 6.8330 & 26.1953 & 19.8255 & 27.3541 & 3\\
5 &	10.5239 & 9.0375 & 25.9106 & 27.2413 & 25.4637 & 10\\
6 & 13.0418 & 6.9784 & 25.6200 & 21.1063 & 27.1561 & 7\\
7 & 12.6973 & \textbf{6.8267} & 25.7065 & 21.8973 & 26.7451 & 1\\
\hline
\end{tabular}
\end{table}

\subsection{Best Parking Lot Recommendation}\label{subsec: Best Parking spot}
Using the formulas discussed in section \ref{subsec: Best Parking spot}, table \ref{tab:distance_and_parking} shows the values of 
the distance $d_{i}$ in Kilometers from 5 driver starting locations and $m_{i}$, the number of parking spots detected at lot $i$. Different values of $alpha$ lead to different solutions to the optimization problem that recommends a parking lot. We look at the solutions for values of $alpha \in { 0.001, 0.01, 0.1, 0.25, 0.5, 0.75, 0.9, 0.999}$, to see how different preferences in distance and abundance of parking spots leads drivers to different parking lots.
In table \ref{tab:Recommendation_results}, parking lot 6 is the recommended parking lot from most starting points when $\alpha$ (short distance preference), is small. This is because parking lot 6 has the most available parking spaces. For moderate values of $\alpha$, parking lot 2 is most preferred for more starting points although it has the fourth highest number of parking spaces. This is largely due to its centrality to most starting points. For higher values of $\alpha$, even lot 1, with only 1 parking space is recommended for one starting point since it is the closest and preference is given to distance over space availability. In practice, the performance of the model in detecting different objects does not pose any foreseeable health or safety concerns as the system is designed only as a recommendation system, and not a tool for autonomous driving. We are particularly happy the detection of cars and car plates is done very well, this is key to being able to mask car plates away in production to preserve peoples write to privacy. With more data, the current model can be improved to perform better detecting humans. We make the model and dataset available on \href{https://huggingface.co/MoseliMotsoehli/DeepGeoPark}{hugging face models}

\begin{table}
\centering
\caption{Recommended parking lot \# based on space and distance optimization for a driver in different starting points and preference for space availability vs distance}\label{tab:Recommendation_results}
\begin{tabular}{|l|l|l|l|l|l|}
\hline
 &\multicolumn{5}{c}{\bfseries Starting Point}\\
 \hline
{\bfseries $\alpha$} & {\bfseries Bushhill} & {\bfseries Waterval Ct} & {\bfseries Dobsonville} & {\bfseries Germiston \textbf{S}} & {\bfseries Eldoraigne} \\
\hline
\hline
$10^-3$  & 6 & 6 & 6 & 6 & 6\\
$10^-2$  & 6 & 6 & 6 & 6 & 6\\
$10^-1$  & 5 & 6 & 6 & 6 & 6\\
0.25 & 5 & 6 & 6 & 6 & 5\\
0.5 & 3 & 4 & 2 & 2 & 5\\
0.75 & 3 & 4 & 5 & 2 & 5\\
0.9 & 3 & 4 & 2 & 2 & 3\\
0.999 & 3 & 7 & 2 & 2 & 3\\
\hline
\end{tabular}
\end{table}

\section{Conclusions and Future Work}\label{sec:Conclusions and Future Work}
To solve the problem of finding the most convenient available free public parking space, we successfully train and test a detection, segmentation, and classification neural network accompanied by a method for distance calculation from an arbitrary starting point to a parking lot in Johannesburg. This work is the first to capture more of the surrounding scene in parking lots by including car, number plate, and people detection and segmentation, while at the same time using image angles that are more realistic in most public parking lots. The other major contribution of this work to parking lot detection is the added recommendation system that incorporates distance from a driver's live geo-location. We achieve great results on the segmentation IoU measure of cars, number plates, and parking spots.
One avenue of further research into this problem is to investigate the implementation details of such systems, do they scale well, and how do multiple requests from multiple drivers in real-time affect the reliability of inferences from the detection system. It is also interesting to think about how a driver's live location as he approaches a recommended parking lot, should affect other users' parking lot recommendations in real-time. This would introduce a need to use a more realistic distance measure such as the Manhattan distance to calculate the exact route distance and time.

\bibliographystyle{unsrtnat}
\bibliography{Public_Parking.bib} 
\end{document}